\documentclass[10pt,twocolumn,letterpaper]{article}

\usepackage[pagenumbers]{wacv} 
\usepackage[accsupp]{axessibility}
\usepackage{graphicx}
\usepackage{amsmath}
\usepackage{amssymb}
\usepackage{booktabs}

\usepackage[pagebackref,breaklinks,colorlinks]{hyperref}

\usepackage[capitalize]{cleveref}
\crefname{section}{Sec.}{Secs.}
\Crefname{section}{Section}{Sections}
\Crefname{table}{Table}{Tables}
\crefname{table}{Tab.}{Tabs.}

\def\wacvPaperID{644} 
\def\confName{WACV}
\def\confYear{2024}

\begin{document}

\title{One Style is All You Need to Generate a Video}

\author{Sandeep Manandhar and Auguste Genovesio\\
IBENS, Ecole Normale Supérieure\\
75005 Paris, France\\
{\tt\small sandeep.manandhar@bio.ens.psl.eu, auguste.genovesio@ens.psl.eu}
}

\twocolumn[{
\maketitle
    
}]

\begin{abstract}
\textcolor{black}{
In this paper, we propose a style-based conditional video generative model. We introduce a novel temporal generator based on a set of learned sinusoidal bases.  Our method learns dynamic representations of various actions that are independent of image content and can be transferred between different actors. Beyond the significant enhancement of video quality compared to prevalent methods, we demonstrate that the disentangled dynamic and content permit their independent manipulation, as well as temporal GAN-inversion to retrieve and transfer a video motion from one content or identity to another without further preprocessing such as landmark points.}
\end{abstract}

\section{Introduction}

Image synthesis has seen significant advancements with the development of generative models. However, generative models of videos have not been as successful, and controlling the dynamic generation process has been a major challenge. This is largely due to the complex spatio-temporal relationships between content/actors and dynamic/actions, which makes it difficult to synthesize and control the dynamics independently. Several methods have been proposed to address this challenge, each with their own design principles. Broadly speaking, there are two primary classes of video generative models: 3D models that learn from 2D+time volumetric data by employing 3D convolutional neural networks (CNNs), and 2D models that generate a sequence of 2D frames while disentangling the spatio-temporal components of a given video distribution. Many of the earlier methods took the former approach treating each video clip as a point in latent space, thus making the manipulation in such space hardly possible. The latter approach is not only more resource-efficient, but also allows for greater control over the generation process, as demonstrated by \cite{coninvert21, DwNet, thinplate22}. However, these methods require some pre processing (optical flow, pose information) to manipulate the generated videos.

In their work, \cite{GrathwohlW16} introduced a variational encoder for visual learning, which assumes that higher-level semantic information within a short video clip can be decomposed into two independent sets: static and dynamic. With similar notion, \cite{DrNET17} employed two separate encoders to produce content and pose feature representations. Pose features are processed by an LSTM to predict future pose information which is then used along with the current content information to generate the next frame. The idea of treating content and motion information independently has laid a foundation for many works in video generation.

Instead of considering a video as a rigid 3D volume, one can model it as a sequence of 2D video frames $x(t)\in \mathbf{R}^{3\times H\times W}$, where $t$ is the temporal point, $(H,W)$ are the height and the width of the video frame. An image generator $G(z)$ can be trained to produce an image $x'\sim x(t)$ from a vector $z$ coming from a latent space $Z \in \mathbf{R}^d$, where $d<H\times W$. However, the problem at hand is to come up with a sequence of $z(t)$ that can be fed into $G(z)$ to produce a realistic video frame sequence. And, if such $z(t)$ can be obtained, how can we manipulate the video generation process?

The authors of \cite{TGAN2017} proposed to first map a latent vector to a series of latent codes using a temporal generator. An image generator would then use the set of codes to output video frames. MOCOGAN, \cite{tulyakov2018mocogan}, on the other hand proposed to decompose the latent space $Z$ into two independent subspaces of content $Z_c$ and motion $Z_m$. $Z_c$ is modeled by the standard Gaussian distribution, whereas $Z_m$ is modeled by a recurrent neural network (RNN). The content code remains the same for a generated video, while motion codes varies for each generated frames. MOCOGAN-HD \cite{tian2021a} and StyleVideoGAN \cite{fox2021stylevideogan} took advantage of a pretrained StyleGAN2 \cite{Karras2019stylegan2} image latent space and proposed to traverse in the latent space using RNNs to produce video frames.

Interestingly, in the context of a pretrained StyleGAN2 network, one can perform GAN inversion\cite{coninvert21} on a image sequence to obtain its latent representation. StyleGAN2 produces a continuous and consistent latent space, where close by latent vectors map to similar realistic images. Taking advantage of this property, the latent vector obtained by optimization from the previous frame can be used as the starting point to search for the latent vector of the next frame, thus optimizing for minor changes. Upon simple linear projection (such as PCA) of the latent trajectory of a movie optimized in such manner, we can observe that the higher components are similar to cosine waves (see Appendix Section 1). The author in \cite{Hess2} also made this observation in the context of protein trajectory simulation, where he finds that the cosine content of the principal components are negatively related to the randomness of the simulation. In the case of optimized vectors corresponding to the inverted images, they are correlated. Hence, the waves are obvious and visible. This hints us that sinusoidal bases could naturally facilitate training of a StyleGAN generator to produce image sequences.

To this end, we propose a temporal style generator in order to generate videos using StyleGAN2's sythesis network. We use a \textit{time2vec} \cite{t2v} network to introduce a temporal embedding from where the temporal styles will be generated. \textit{time2vec} network provides a learnable Fourier bases. By scaling the Fourier bases using a single motion style vector, we propose to produce diverse and arbitrary length videos.
Main contributions of our work are as follow:
\begin{itemize}
    \item We integrate a novel temporal latent space in StyleGAN’s generator network using a sinusoid-based temporal embedding. 
    \item We evaluate our method against prevalent methods in an unconditional setting, demonstrating a significant enhancement of video quality. 
    \item We propose several approaches to rigorously evaluate conditional video generation through contexts such as talking faces and human activities.
    \item \textcolor{black}{We recover motion from real input videos and map it to our learned latent motion style space via GAN-inversion. This further facilitates the manipulation of temporal style of the generated videos.}
\end{itemize}

We trained our model on videos of talking head (MEAD \cite{MEAD20}, RAVDESS \cite{RAV}) and human activities (UTD-MHAD \cite{UTD15}). Besides the Fréchet video distance (FVD) \cite{FVD18} metric, we conducted human evaluation focused on the realism of the generated videos using the MEAD dataset. Additionally we proposed LiA (Lips Area) metric to evaluate the videos generated from the MEAD dataset. We also benchmarked our results using publicly available method for human action recognition with UTD-MHAD dataset.

\section{Related work}
The domain of video synthesis consists of tasks such as future frame prediction \cite{Finn16, DMS16,WalkerDGH16,DrNET17}, frame interpolation \cite{Niklaus_CVPR_2017,slomo18, xiang2020zooming} and in our context, video generation from scratch\cite{VGAN16}. Video generation follows the success of image generative adversarial models which can produce highly controllable images of remarkable quality \cite{GAN}. Much focus has been given to temporal extension of such GANs. \cite{tulyakov2018mocogan,TGAN2017,TGAN2020, TSGAN21_WACV} have adopted the strategy to use content and motion codes by leveraging on 2D image generator. MOCOGAN-HD\cite{tulyakov2018mocogan} used a pretrained StyleGAN2's network\cite{Karras2019stylegan2} and trained a RNN model to simply explore along the principal components of the latent space. Recently, \cite{LVGAN2022} also proposed a style-based temporal encoding for a 3D version of StyleGAN3's synthesis network\cite{Karras2021} where temporal codes are generated by a noise vector filtered by a fixed set of temporal low pass-filters. \cite{Digan2022} used implicit neural representation (INR) \cite{chen2022vinr, Chen_2021_CVPR} to model videos as continuous signal. Finally, StyleGAN-V \cite{Skorokhodov_2022_CVPR}, relied on training a modified StyleGAN2 generator with an INR-inspired positional embedding for the successive video frames. Both of these methods produce videos with arbitrary frame rates. \textcolor{black}{Our method is related to StyleGAN-V as it uses StyleGAN2 synthesis network. However, while StyleGAN-V requires multiple random input vectors to obtain a single trajectory, our approach requires only one such input vector. The latter allows us to manipulate the temporal aspect of the generated videos \cite{coninvert21,roich2021pivotal}}.

Conditional generative models are another exciting field of research. Besides explicit vector based labels, text, audio and images have been used in conditioning for frame generation. \cite{ImaGin20} proposes a simple and efficient 3D CNN based generator that takes a single image and a conditioning label as an input to generate videos. \cite{x2face} takes a source frame with one human face and generates video that has pose and expression of another face in a driving video. \cite{wang2018vid2vid} conditioned their video generation on semantic maps where objects present in the frame are labelled with colors. The network can also take information like optical flow and pose information during the training. \cite{Songsriin2019FaceVG} generated videos of talking face using sequence of facial landmarks of target face. \cite{thinplate22} is yet another image-conditioned video generation model, which has dedicated networks for motion prediction and keypoint detection. However, it is not straightforward to generated videos with arbitrary frame rates with  image-conditioned models. 
\textcolor{black}{
Recently diffusion based models have been employed to generate videos as they can output high quality images and demonstrated great flexibility while used with language based prompts. \cite{ho2022video} proposed a 3D U-Net based diffusion model for text-to-video generation. Following this \cite{singer2023makeavideo} proposed another text-to-video generation method that makes use of efficient 3D convolutions and temporal attention modules. They also added an embedding in order to specify the frame rates. The authors of \cite{alignlat} introduced a temporal dimension to the latent space of a pretrained text-to-image diffusion model to generate videos. The work in \cite{PVDM} introduces a video encoder that projects a video clip to a 2D latent representation, which is further processed by a diffusion model to synthesize videos. However, diffusion models are notorious for being resource hungry and slow due to their gradual iterative denoising process at training and inference times. In our study, we have limited our comparative study to GAN-based models only.}
\section{Method}

Our method contains two main components: (1) a temporal style generator that drives StyleGAN2's synthesis network to produce frames in time-conditioned manner, (2) two discriminators to impose content consistency and temporal consistency. Our generator is further conditioned on actor identity and action classes, though it can be used in unconditional setting.

\subsection{Generator}
\textcolor{black}{
Our generator consists of three distinct networks: a synthesis network $\mathbf{G}$, a temporal style generator $\mathbf{F_t}$, and a conditional embedding $\mathbf{F_c}$ as shown in Fig. \ref{fig:network}.  The synthesis network $\mathbf{G}$, which is based on StyleGAN2, is inherently agnostic to temporal cues when generating images. To ensure the generation of temporally coherent video frames, we introduce a specific temporal embedding, which interfaces with $\mathbf{G}$ to guide the synthesis process using a latent trajectory. This trajectory is derived from the network $\mathbf{F_t}$, which comprises a 4-layer Perceptron (MLP) that maps a random vector $z_m$ to a $k$-dimensional motion style vector $m$. At this stage, the vector $m$ does not encompass any temporal context. An auxiliary network \textit{time2vec} produces sinusoidal bases that are scaled by $m$ to finally output the temporal style vectors $w^t_m$.}\\
\textcolor{black}{
\indent{\textbf{Time2vec}}: Our proposed $k$ dimensional time embedding consists of $k-1$ sinusoidal bases and a linear term as seen in Eq. \ref{eq:t2v}, where the parameters $w_j$ and $\phi_j$ are trainable. 
\begin{equation}
 v_j(t)= \mathcal{F}(\omega_jt+\phi_j),
\label{eq:t2v}
\end{equation}
where $\mathcal{F}$ is the identity function when $j=0$ and the sine function for $1 \leq j \leq k-1$. The linear term $v_0(t)$ represents the time direction. The time $t$ does not need to be discrete as the \textit{time2vec} embedding is continuous. This allows us to generate videos with arbitrary frame rates. However, during the training we use integer valued time-points. We note that StyleGAN-V's time representation lacks the linear term, which might explain why its generation is plagued by unnatural repetitive motion despite its elaborate interpolation scheme. By restricting the dynamics to a fixed set of sinusoidal functions, we avoid over-fitting to the training data, and make the model more robust and generalizable to unseen data. Moreover, since sinusoidal functions are periodic, they can naturally capture cyclic patterns in the data (e.g. lip movement, hand waving).
We obtain a temporal style vector as an input to $\mathbf{G}$ using the following set of equations:
\begin{align}
    m &= \mathbf{F_t}(z_m),\\
    w^t_m &= m*v(t),\\
    w^{t+1}_m &= m*v(t+1).\label{eq:1} \end{align}
The product of motion style $m$ and temporal embedding vector $v(t)$ is a temporal style vector $w^t_m$. Note that a single $m$ is used to compute temporal styles for consecutive frames. 
Additionally, $\mathbf{F_c}$ encodes the action and actor embeddings and outputs a content style vector $w_c$. It defines the general appearance of the actor along with the nature of the action. To generate a frame at time $t$, both temporal and content styles $[w_c, w^t_m]$ are concatenated and injected to the synthesis blocks. During the training, we generate three consecutive frames for each video element of the batch. The triplets share the same vector $m$ while their temporal embeddings are generated from their respective time points. During the inference, a single vector $m$ is enough to generate a long duration video.  We leverage this ability of $m$ to encapsulate the entire dynamics of a sequence to compute a temporal GAN inversion, as described in Section \ref{ssec:GAN}. 
}
A basic structure of the generator network is shown in Figure \ref{fig:network}. 
To ensure the smooth integration of action-id embeddings, we employ a ramp function \cite{collpaseCon} that linearly scales the vectors derived from the action-id embedding with a factor ranging from 0 to 1, in a scheduled manner.

Unlike StyleGAN-V, we choose to stay closer to the StyleGAN's original principle, which is to allow variations in input only through the style vectors. Furthermore, our time embedding fundamentally differs from StyleGAN-V's in its design. StyleGAN-V requires multiple randomly sampled vectors to compute wave parameters which ultimately defines the motion of a generated video. In contrast, our wave parameters are independently learned and are fixed during inference. Our latent vector $m$ interacts with the waves only as an amplitude scaling factor. Hence, our time representation is simpler and manipulable. We leverage these advantages in Section \ref{ssec:GAN} to perform GAN-inversion of the motion style using off-the-self methods, which cannot be achieved with StyleGAN-V.
\textcolor{black}{
\begin{figure*}[t]
\begin{center}
   \includegraphics[width=0.96\linewidth]{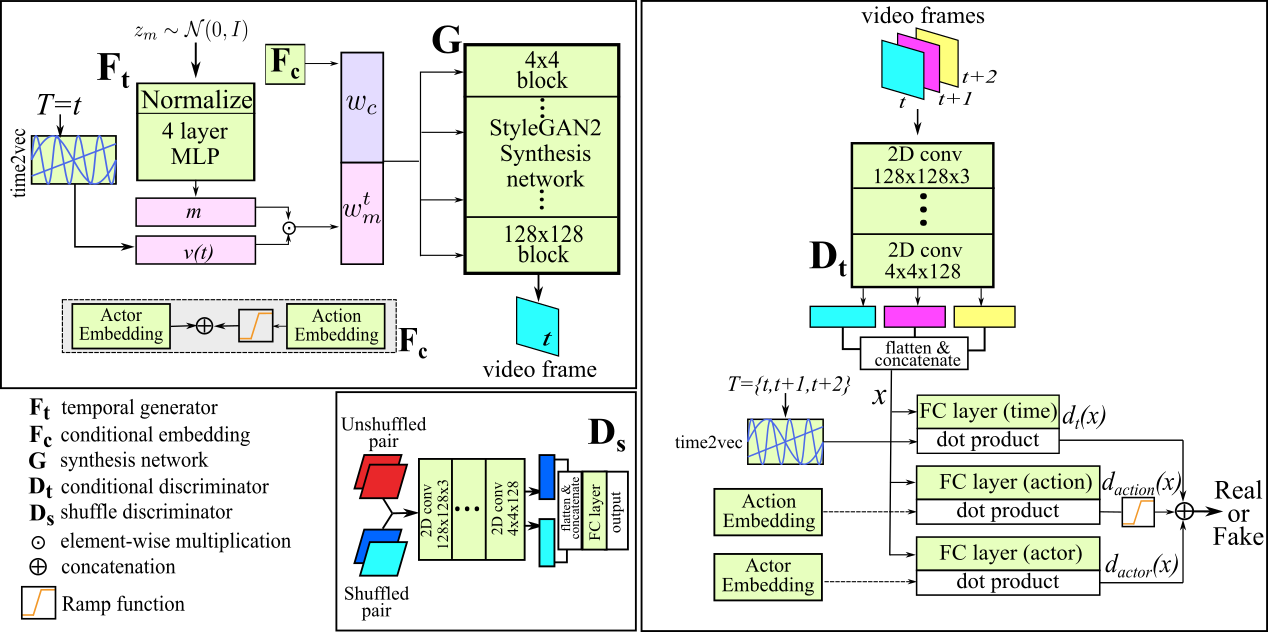}
\end{center}
   \caption{\textcolor{black}{Proposed model: a temporal style generator $\text{F}_\text{t}$ equipped with a \textit{time2vec} module generates the motion code. $\text{F}_\text{c}$ outputs a vector formed by concatenation of actor and action embeddings. Similar embeddings are activated in $\mathbf{D_t}$'s final layer. A ramp function\cite{collpaseCon}, which gradually increases from 0 to 1, is used to scale the action embedding vectors in both $\text{F}_\text{c}$ and $\mathbf{D_t}$. Here, $\text{G}$ is the StyleGAN2's synthesis block.}}
\label{fig:network}
\end{figure*}
}

\subsection{Discriminators}
\indent \textbf{Shuffle discriminator}: Consistency in content over time is a crucial aspect of video generation. Although the $\textit{time2vec}$ module in $\mathbf{G}$ provides temporal bases to guide motion learning, it does not ensure consistency in content across the sequence. In order to address this, we design a 2D-CNN based discriminator $\mathbf{D_s}$ (as seen in Figure \ref{fig:network}) that evaluates whether the frame features are consistent or not. During the training of $\mathbf{D_s}$, each batch element consists of two frames. For the fake adversarial example, pairs of frames are shuffled among the batch to contain two different contents. In contrast, for the real example, the pairs are consecutive frames drawn from real videos. The feature maps of the pairs undergo a series of 2D convolutions, are flattened, and then concatenated into a single vector before passing through a fully connected layer. During the training of $\mathbf{G}$, a batch of unshuffled fake pairs is input to $\mathbf{D_s}$.

\begin{figure*}[t]
\begin{center}
   \includegraphics[width=0.95\linewidth]{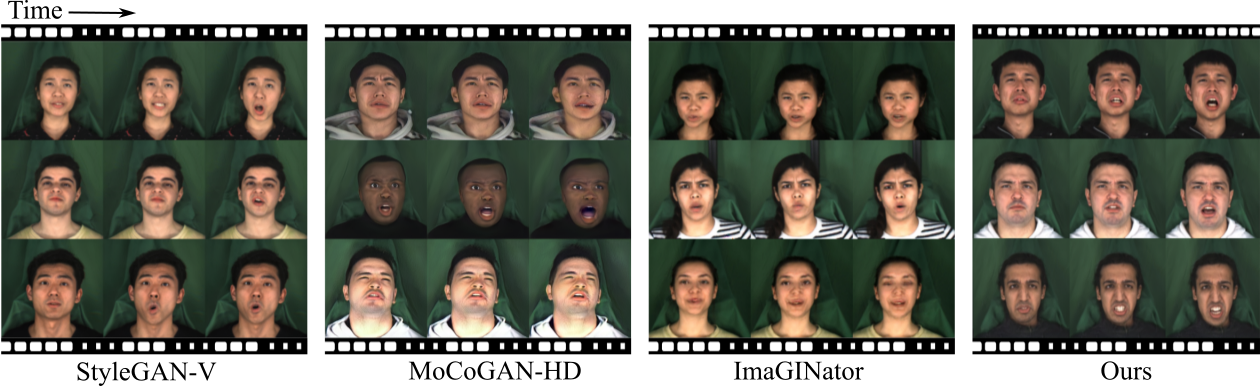}
\end{center}
   \caption{Here, we show few samples of generated frames by different methods. Please refer to the accompanying supplementary videos for more examples. Note that the generation of StyleGAN-V lacks natural facial motion. MOCOGAN-HD and ImaGINator's generated videos are ridden with artefacts. Meanwhile, our generation does not have these issues.}
\label{fig:all_mead}
\end{figure*}

\indent \textbf{Conditional discriminator}: To ensure temporal consistency in the generated videos, we adopt a time-conditioned discriminator, inspired by prior works of \cite{miyato2018cgans, Digan2022}. The discriminator, denoted as $\mathbf{D_t}$, takes in a batch of video triplets (three consecutive frames per video) along with their respective time information, and learns to distinguish real videos from fake ones based on their temporal coherence. Then the video frames are processed by a set of 2D CNNs and a linear layer $d_{t}(.)$ to produce frame features. These features are then concatenated following the temporal order. $\mathbf{D_t}$ is equipped with another \textit{time2vec} module which enforces learning of a time representation. The temporal encoding for the three input time points are also concatenated. The dot product of these concatenated vectors is computed to generate the final score\cite{miyato2018cgans}. \textcolor{black}{Design-wise, $\mathbf{D_t}$ is similar to StyleGAN-V's discriminator as it was also inspired by the aforementioned works. However, our $\mathbf{D_t}$ learns the temporal order using absolute time information via \textit{time2vec} in contrast to time difference conditioning used in StyleGAN-V. The use of absolute time information increases flexibility as it allows $\mathbf{D_t}$ to evaluate an arbitrary number of frames. We demonstrate this in our ablation studies where we use a single frame instead of three time frames.}

As shown in Figure \ref{fig:network}, two additional linear layers ($d_{action}(.), d_{actor}(.)$) are present at the level of $d_{t}(.)$, which produce actor and action representations. Dot products are computed between the corresponding embedded vector and the feature vector. The final output of the discriminator is the weighted sum of the three dot products. We use the same ramp-up function to scale $d_{action}(.)$ as in the generator \cite{collpaseCon}.

\section{Experimental settings}
We focus on the conditional generation of videos. However, we also present the results of unconditional generation for comparative studies.
\subsection{Datasets}
We have used three publicly available video datasets with their labels: MEAD \cite{MEAD20}, RAVDESS \cite{RAV} and UTD-MHAD \cite{UTD15}. Our MEAD training set contains 30 individuals talking while expressing 8 different emotions ($18883$ videos). We train our network only with the sequences where generic sentences are being recited. We set aside the emotion specific dialogues as unseen test sequences. For the training, we chose $128\times 128$ image dimension and between $60-170$ frames as the dataset contains videos of variable length.

The RAVDESS dataset contains 24 talking faces also with 8 different emotions (not same categories as MEAD). To create a test set, we exclude sequences of 7 different emotions for four individuals. Though the dataset set contains only two dialogues, compared to over 20 dialogues in MEAD, RAVDESS contains more variation in head movements of the actors.

UTD-MHAD contains $754$ videos of 8 individuals performing 27 different actions. The video frame size is $128\times128$ with variable video length ($33-81$) as provided in the dataset. We created a test set by excluding videos of each action sequence performed by few selected target actor from the training set. Thus, we train the network to learn motion and content independently.

\subsection{Baseline Methods}
For the conditional video generation, we choose ImaGINator \cite{ImaGin20} as our baseline. Though it requires a conditional input image to generate videos, it is free of any additional representation like pose or motion maps. We adapted its network to output $128\times128\times32$ size image (originally $64\times64\times32$). We trained it on MEAD and UTD-MHAD datasets for up to $5$K epochs.

To demonstrate that our generator does not falter in video quality, we choose MOCOGAN-HD \cite{tulyakov2018mocogan} and StyleGAN-V \cite{Skorokhodov_2022_CVPR} as our baselines in unconditional setting as they both use StyleGAN2's image synthesizer. For MOCOGAN-HD, we first trained a StyleGAN2 network on MEAD dataset with $256^2$ image size for upto 150K iterations. Then the MOCOGAN-HD network was trained with the hyper-parameters set as suggested in the author's implementation. For StyleGAN-V, we trained on  with image of dimension $256^2$, with a batch size of 64 and with up to 25000K images according to the author's implementation. 

\subsection{Training}
We trained our method on a single Nvidia's A100 GPU with $80$GB VRAM. The training image size was $128^2$ with a batch size of $16$ triplet frames. The hyperparameters for the generator, discriminators and the optimizers were kept the same as suggested in \cite{Karras2019stylegan2}. The transition factor $\lambda$ of action-id vectors in both generator and discriminator started at $4000$ iterations and ended at $6000$ iterations, which was set empirically. We trained our model on all datasets for up to $120k$ iterations which took about 2 weeks. Our method can generate longer videos with diverse motion types and arbitrary frame rates. 

\section{Results}
\subsection{Video quality is improved}
Table \ref{tab:fvd} reports the FVD scores of the generated videos by all the methods. Our conditional method (Ours(C)) scores the best which is in agreement with the videos provided in the supplementary data. Few frames of the generated video samples are depicted in Figure \ref{fig:all_mead}. Motion artifacts are strongly present in MOCOGAN-HD and ImaGINator's output. Though StyleGAN-V generates long duration videos, it suffers from erratic, repeated motion. Our methods (both conditional Ours(C) and unconditional Ours(UC)) produce far better results. \textcolor{black}{We have reported FVD score computed over 64 frames only for StyleGAN-V and our method as other baselines are incapable of long duration video generation.}

To assess the preservation of the actor's identity, we computed the ArcFace\cite{arcface} similarity between the frames of the generated videos. ArcFace computes the cosine similarity between the feature vector of the first and the successive frames obtained from face recognition network. As seen in Table \ref{tab:fvd}, our methods preserve the appearance of the actor throughout the sequence while MOCOGAN-HD is not consistent generating the same face over the sequence. The authors of \cite{Skorokhodov_2022_CVPR} also made this observation. 

\begin{figure}[]
\begin{center}
   \includegraphics[width=0.8\linewidth]{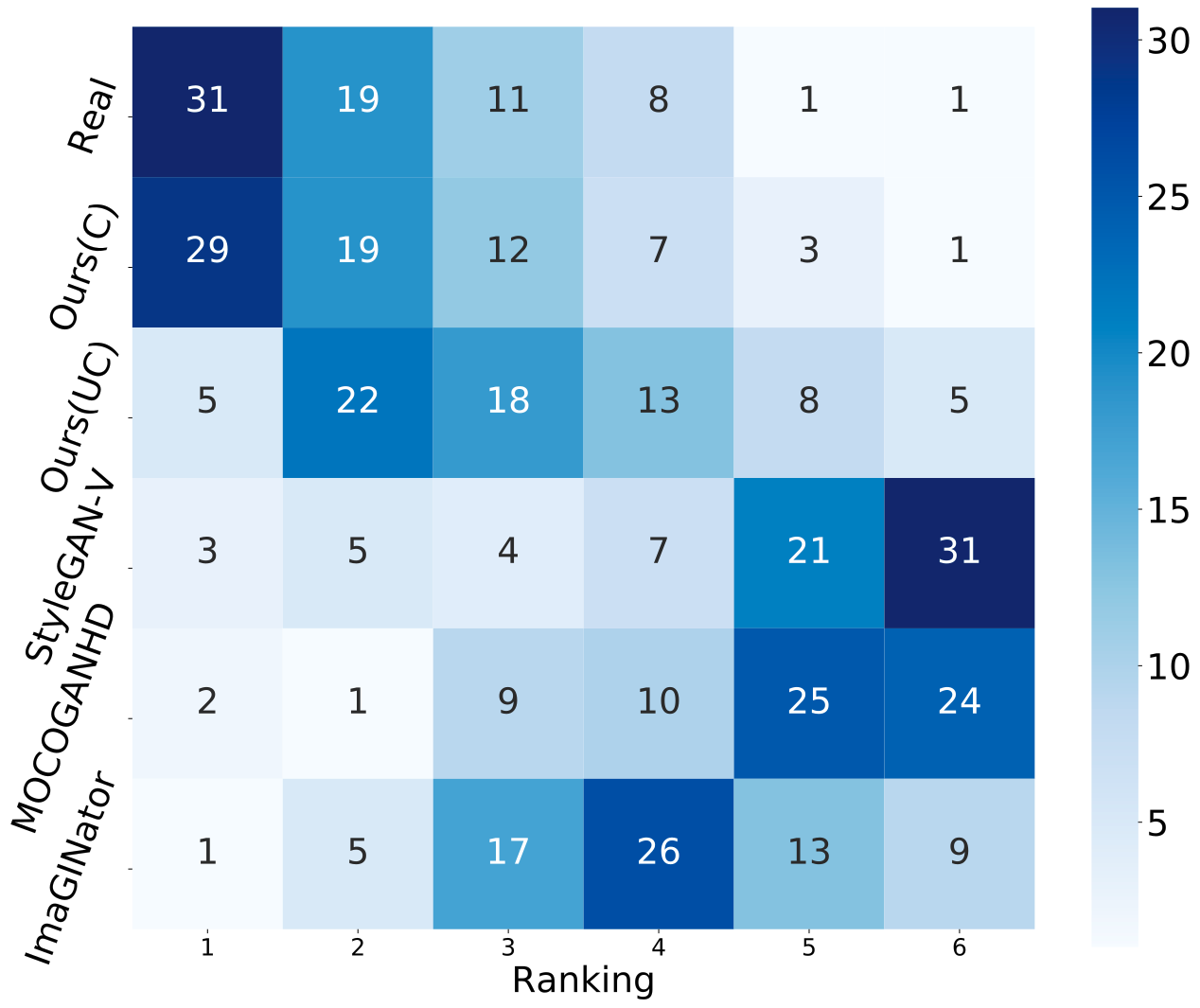}
\end{center}
   \caption{Human preference ranking for different videos. Videos generated by our conditional model (Ours(C)) tops the preference over other methods.}
\label{fig:ranks}
\vspace{-2pt}
\end{figure}

The FVD score is widely used to evaluate video quality. However, as it is a comparison of distributions of representations in a high dimensional space, it may not accurately characterize the true quality of the video. The same can be said about the ArcFace score. Furthermore, these metrics can be influenced by factors such as spatial resolution, video length, etc. To complement these metrics, a human evaluation was conducted to assess the realism of the generated videos. To conduct the human evaluation, we generated 10 sets of videos, each consisting of 6 videos with 32 frames (1 real video and 5 generated videos using the proposed methods and the baselines). We asked 25 university students and researchers to watch 3 randomly selected sets and rank the 6 videos based on their perceived realism. The ranking distributions of the survey is presented in Figure \ref{fig:ranks}. Notably, videos generated with Ours(C) and Ours(UC) models consistently ranked higher than those generated using the baseline methods. This demonstrates that our method produces more realistic videos compared to existing approaches.
\begin{table}[t]
\centering
\resizebox{0.38\textwidth}{!}{%
\begin{tabular}{|c|c|c|l|}
\hline
\textbf{Method}                  & $\text{FVD}_{16/64}{}\downarrow$             & $\bar{\text{r}}_{t}{}\uparrow$ & ArcFace ${\uparrow}$               \\ \hline
ImaGINator                       & 319                           & 0.041                & 0.93$\pm$0.03            \\ \hline
MOCOGAN-HD                       & 272                           & 0.52                 & 0.80$\pm$0.13            \\ \hline
StyleGANV                        & 191/920                           & 0.77                 & 0.92$\pm$0.05            \\ \hline
Ours(UC)                       & 140                           & \textbf{0.79}        & \textbf{0.97}$\pm$0.018  \\ \hline
Ours(C)                        & \textbf{115/655}                 & 0.7                  & 0.96$\pm$0.02            \\ \hline
\end{tabular}
}
\caption{All the scores pertain to the training with MEAD dataset. $\text{FVD}_{16/64}$ is computed with 16 and 64 (only for StyleGAN-V and ours(C)) frames. $\bar{\text{r}}_{t}$ is the average correlation coefficient of the LiA signals. ArcFace is the average of the cosine similarity between the features of the first frame and the successive frames.}
\label{tab:fvd}
\end{table}

\subsection{Temporal style encodes temporal semantic}
While we demonstrated that the video quality is improved, the aforementioned metric cannot assess the preservation of temporal semantics across different sequences. We then propose a new metric named LiA (for Lips Area) to evaluate our ability to reproduce the semantic of talking-face videos while changing the content such as the actor-id or action-id (emotion). LiA value computes the polygonal area of the lips detected using face landmark detectors \cite{DlibmlAM}. A LiA signal is then obtained by computing LiA value sequentially for each frames of a generated or a real video. Though there are other factors such as eye brows and head orientation that contribute to the overall dynamics of a talking face, we focus on lip motion as it appears to be the most dynamic part of the face on this dataset.
We generated $100$ different sequences using different content styles and the same temporal style for the baseline methods. The average correlation coefficient $\bar{\text{r}}_t$ of the LiA signals of the generated videos by all the methods are reported in Table \ref{tab:fvd}. We observed that even for the same temporal style, the ImaGINator produced different motion pattern depending on the starting input frame. MOCOGAN-HD has relatively low $\bar{\text{r}}_t$ score as the face unusually distorts over time.



\begin{table}[]
\centering
\resizebox{0.4\textwidth}{!}{%
\begin{tabular}{|l|l|l|l|}
\hline
Methods & StyleGAN-V & ImaGINator & Ours(C) \\ \hline
$\text{FVD}_{16}{}\downarrow$          & 421.32     & 649.23     & \textbf{184.55}  \\ \hline
(top-1, top-3)\% $\uparrow$         & n/a     & (0,25)     & \textbf{(68.5, 93.5)}  \\ \hline
\end{tabular}
}
\caption{FVD score for the UTD-MHAD dataset. Though we train StyleGAN-V unconditionally, it serves as a good baseline for assessing the video quality. We also report top-(1,3) action recognition accuracies among 27 action classes.}
\label{tab:fvd_utd}
\vspace{-5pt}
\end{table}

\begin{figure*}[h]
\centering
\begin{subfigure}{.5\textwidth}
  \centering
  \includegraphics[width=.8\linewidth]{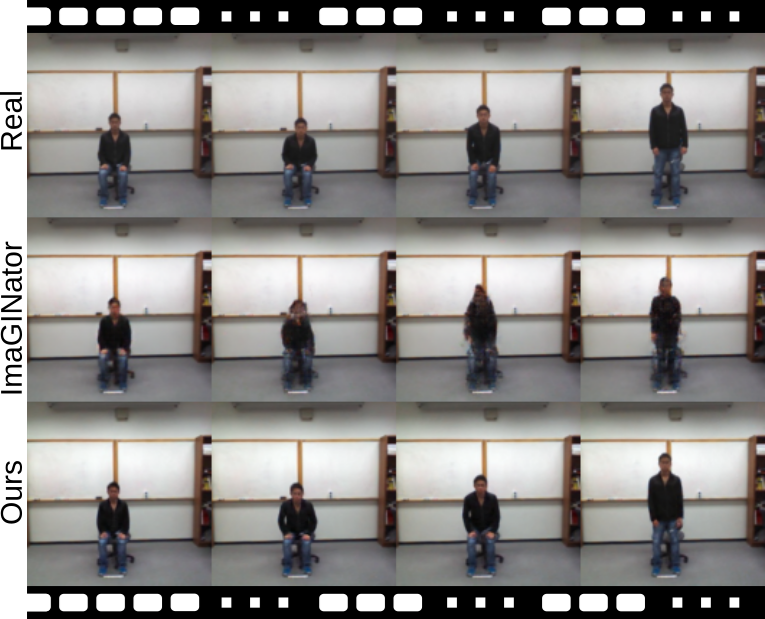}
  \caption{Sit-to-Stand sequence}
  \label{fig:sub1}
\end{subfigure}%
\begin{subfigure}{.5\textwidth}
  \centering
  \includegraphics[width=.8\linewidth]{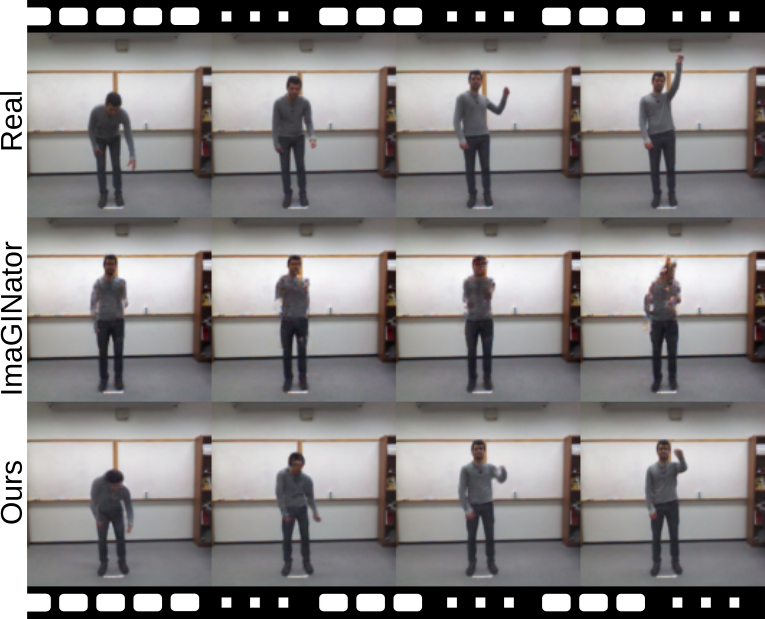}
  \caption{Pick-and-throw sequence}
  \label{fig:sub2}
\end{subfigure}

\caption{Dynamics transfer in unseen conditions. Both panels display few frames sampled from videos of real sequence (top), generated sequence by ImaGINator (mid), and our method (bottom). The actors were never seen performing these actions (top) during the training.}
\label{fig:utd_dyntransfer}
\end{figure*}
\begin{figure*}[h]
\begin{subfigure}{.5\textwidth}

  \centering
  \includegraphics[width=.85\linewidth]{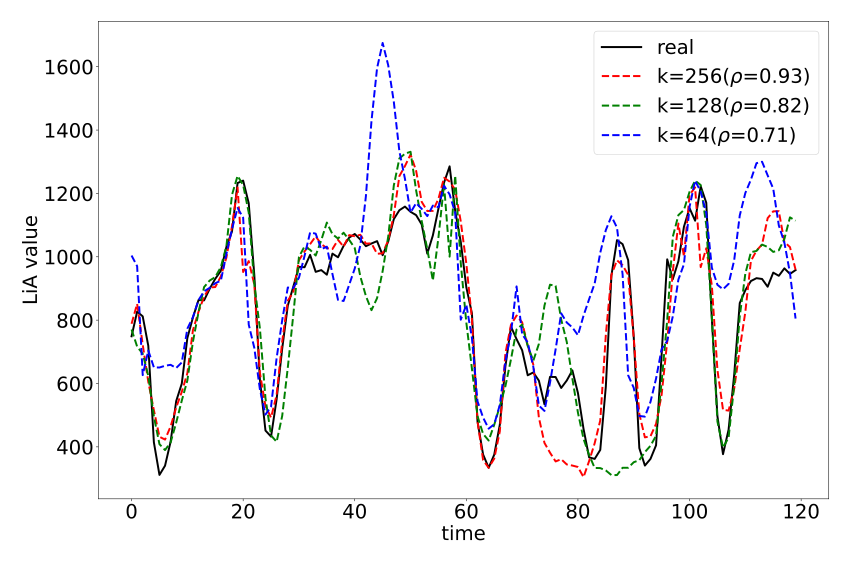}
  \caption{LiA signals of real, and inverted sequences}
  \label{fig:liasub1}
\end{subfigure}%
\begin{subfigure}{.5\textwidth}
  \centering
  \includegraphics[width=.85\linewidth]{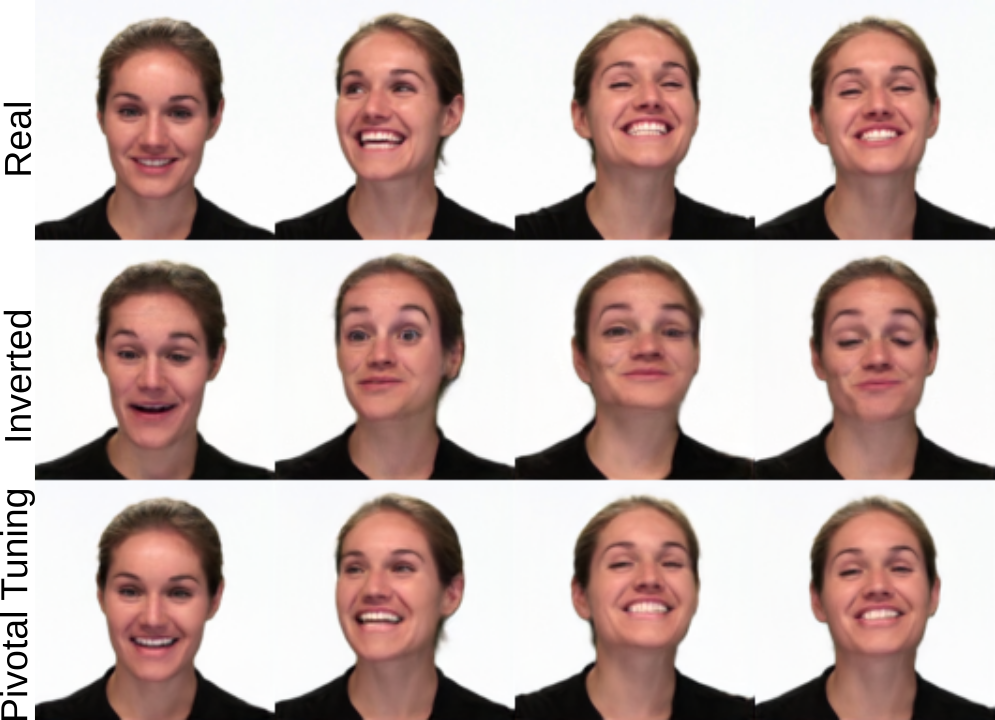}
  \caption{GAN-inversion and Pivotal tuning}
  \label{fig:invsub2}
\end{subfigure}
\caption{\textcolor{black}{(a) The correlation coefficients ($\rho$) between LiA signals of real and inverted videos suggest that the network with a higher number of sinusoidal bases generates more faithful videos. (b) Pivotal tuning further improves the facial structure even though most of the motion is already recovered in the first step.}}
\label{fig:h2e}
\end{figure*}
\subsection{Generation of unseen coupled conditions}
We generate videos of unseen actor-action combination only present in the test set. Figure \ref{fig:utd_dyntransfer} shows a few selected frames of real videos, and generated videos by ImaGINator, and our conditional method. Our method is able to successfully transfer a learnt action to an actor who was never seen performing this action in the training set.
To evaluate our method in this dataset, we additionally train an action recognition model using the implementation of \cite{posec3d}. We train the model on skeletal key points extracted from the video frames of our training set which contains 27 different actions. The trained model was able to achieve (77\%, 100\%) top-1 and top-3 accuracies on the real test cases. In our generated case, it was able to achieve (68.5\%, 93.5\%) top-1 and top-3 accuracies. We present the confusion matrices for 27 different classes in the supplementary data (in Appendix Section 4). Our model not only generated high-quality videos, as shown in Table \ref{tab:fvd_utd}, but also accurately captured many actions. On the other hand, ImaGINator performed poorly on this dataset, with evidence of mode collapse in the type of motion despite the conditioning during inference. We have included the generated videos in the supplementary data.

\textcolor{black}{
\subsection{Motion recovery with GAN inversion} \label{ssec:GAN}
A talking face video can be generated with a random temporal style, however it is a challenging task to map a real motion to a learned temporal representation. However, thanks to our simple temporal representation $m$, it is possible to extract a temporal style of real lip movement, without the need of any motion computation or landmark point detection, directly by GAN-inversion. To the best of our knowledge, it is the first time a GAN-inversion for temporal styles is proposed to recover a reusable dynamic representation.\\
In the following experiments, we invert the temporal style of unseen videos from test cases of MEAD and RAVDESS dataset. For the MEAD dataset, we recover the motion from real video of actors pronouncing sentences which were excluded from the training set. We assume that the excluded dialogues carry unseen lip motions, and recovering such motion should demonstrate the flexibility of our temporal representation. In these experiments, conditional labels are set to the known actor and emotion of the real input sequence and only $m$ is recovered by optimization. To this end, we minimize the sum of the LPIPS loss \cite{zhang2018perceptual} and the MSE between $N$ real and generated frames: 
\begin{align}
m^* &= \underset{m}{\mathrm{arg\,min}}\sum_{t=0}^N \mathcal{L}(I(t), G([w_c, w^t_m])),
\end{align}
where, $I(t)$ is the real video frame at time $t$, $\mathcal{L}$ is the summation of the two losses, and $m^*$ is the optimized motion style vector.
Figure \ref{fig:liasub1} shows an example of LiA signals for real and inverted videos using our model trained with $k=64, 128, 256$. The higher the number of sinusoidal bases used in the generator, the more faithful the recovered motion is to the real video. We performed the inversion and LiA signal analysis for $39$ different emotion specific sentences excluded from the training set (see Appendix Section 6 for the complete list), and report the average correlation to be $0.6$, $0.79$ and $0.91$ for $k=64, 128, 256$ respectively. The evaluation was done for input videos with $120$ frames because of the memory limitation. This implies that a single vector $m^*$ can faithfully represent the dynamic of at least up to $120$ frames. Furthermore, in Figure \ref{fig:invsub2}, the inversion is able to recover the large movement of head in the RAVDESS dataset. The facial structure is further improved using pivotal tuning \cite{roich2021pivotal} where we adjust generator's weight by fixing the previously optimized $m^*$. Thus the recovered motion in the form of $m^*$ can then be transferred to another actor. We believe this is a novel way for re-enactment between different actors and actions.}

\begin{table}[]
\centering
\resizebox{0.4\textwidth}{!}{%
\begin{tabular}{|l|l|l|}
\hline
\textbf{Ablation}  & $\text{FVD}_{16}$ & \begin{tabular}[c]{@{}l@{}}accuracy (top-1/top-3)\%\end{tabular} \\ \hline
$\mathbf{D_t}$(1 time-point) & \textbf{179.73}  & 59.2/79.6                                                                             \\ \hline
$\mathbf{D_t}$(3 time-points) & 184.55  & \textbf{68.5/93.5}                                                                                \\ \hline
w/o $\mathbf{D_t}$           & 526.21  & 42.6/80.5                                                                                  \\ \hline
\end{tabular}
}
\caption{FVD and classification accuracy for UTD-MHAD with three different versions of our model.}
\label{tab:utd_fvd_class}
\end{table}

\subsection{Interpolating conditions over time}
Because our content and motion spaces are highly disentangled, it is possible to edit the attributes of the videos over time. We choreograph a sequence where actors change their expression over time by a linear interpolation in the action embedding space (see supplementary videos). The interpolation does not interfere with the general motion.

\section{Ablation}
\textcolor{black}{
In our ablation studies, we investigate the impact of different components on the performance of our model. Using a higher number of sinusoidal bases improves the recovered motion with GAN-inversion as discussed in the previous section. However, higher number of $k$ leads to small intermittent motion artefacts of eyes in MEAD dataset. For $k=128$, most of the artefacts are unnoticeable.}
We trained $\mathbf{D_t}$ using only one time point instead of three time points, which resulted in a decrease in action recognition accuracy from $68\%$ to $57\%$ for unseen conditions. Secondly, we removed $\mathbf{D_s}$ in the training on the MEAD dataset, which led to an FVD score of $600$. We report the affect of tweaking of $\mathbf{D_t}$ on UTD-MHAD dataset in Table \ref{tab:utd_fvd_class}. We also examined the effect of using a ramp function to schedule scaling of the action-id vectors. We found that without the ramp function, introducing the action-id at the beginning of the training caused the generator to favor one class over the other, while using the ramp function stabilized the quality of the videos for all classes.
\textcolor{black}{
\section{Limitations and Future work}
While our study has achieved convincing and promising results in the realm of style-based conditional video generation and video GAN inversion, several limitations and avenues for future research warrant consideration. First, it is important to note that our experiments are primarily concentrated on scenarios involving single actors executing simple actions. The current method could encounter challenges when attempting to generate video scenes featuring multiple actors with intricate interactions. The empirical choice of $k$, i.e. the number of Fourier bases in our experiments may not be optimal to capture complex dynamics. A possible solution could consist of adopting a multi-resolution approach, whereby lower-frequency bases are introduced during coarser stages, progressively incorporating higher-frequency elements in finer stages. Furthermore, our current video GAN-inversion succeeds in a conditional setting. Without providning the actor-id, the optimization methods fail so far. This model would benefit a robust optimization method that could disentangle the actor from the action during the inversion process. 
}
\section{Conclusion}
In this study, we proposed a video generation model which produces high quality videos in both conditional and unconditional settings. Through  various experiments, we show that the temporal style can independently encode the dynamics of the training data and can be transferred to unseen targets. We demonstrated that it is possible to generate different types of action with high accuracy as seen in UTD-MHAD videos. Our generator produces videos with better fidelity than the prevalent style-based video generation methods as shown by various metrics as well as human preference score. \textcolor{black}{We demonstrate that our method can recover motion of real input videos via GAN-inversion and can faithfully encode the motion of at least $120$ frames with a single temporal style vector. A Pytorch implementation of this work can be found on our project webpage at \href{https://sandman002.github.io/CTSVG}{sandman002.github.io/CTSVG}.} 
\section{Acknowledgement}
This work has received support under the program \textit{Investissements d’Avenir} launched by the French Government and implemented by the ANR, with the references: ANR-10-LABX-54 MEMO LIFE ANR-11-IDEX-0001-02 PSL*. Sandeep Manandhar was funded by Inserm ITMO Cancer - TOTEM. This work was granted access to the HPC resources of IDRIS under the allocation 2020-AD011011495 made by GENCI. 

{\small
\bibliographystyle{ieee_fullname}
\bibliography{egbib}

\begin{thebibliography}{10}\itemsep=-1pt

\bibitem{alignlat}
Andreas Blattmann, Robin Rombach, Huan Ling, Tim Dockhorn, Seung~Wook Kim,
  Sanja Fidler, and Karsten Kreis.
\newblock Align your latents: High-resolution video synthesis with latent
  diffusion models.
\newblock In {\em CVPR}, 2023.

\bibitem{LVGAN2022}
Tim Brooks, Janne Hellsten, Miika Aittala, Ting-Chun Wang, Timo Aila, Jaakko
  Lehtinen, Ming-Yu Liu, Alexei~A. Efros, and Tero Karras.
\newblock Generating long videos of dynamic scenes.
\newblock In {\em NeurIPS}, 2022.

\bibitem{UTD15}
Chen Chen, Jafari Roozbeh, and Kehtarnavaz Nasser.
\newblock Utd-mhad: A multimodal dataset for human action recognition utilizing
  a depth camera and a wearable inertial sensor.
\newblock In {\em ICIP}, 2015.

\bibitem{Chen_2021_CVPR}
Yinbo Chen, Sifei Liu, and Xiaolong Wang.
\newblock Learning continuous image representation with local implicit image
  function.
\newblock In {\em CVPR}, June 2021.

\bibitem{chen2022vinr}
Zeyuan Chen, Yinbo Chen, Jingwen Liu, Xingqian Xu, Vidit Goel, Zhangyang Wang,
  Humphrey Shi, and Xiaolong Wang.
\newblock Videoinr: Learning video implicit neural representation
  for\\continuous space-time super-resolution.
\newblock {\em CVPR}, 2022.

\bibitem{arcface}
Jiankang Deng, Jia Guo, Niannan Xue, and Stefanos Zafeiriou.
\newblock Arcface: Additive angular margin loss for deep face recognition.
\newblock In {\em CVPR}, 2019.

\bibitem{DrNET17}
Emily~L Denton and vighnesh Birodkar.
\newblock Unsupervised learning of disentangled representations from video.
\newblock In I. Guyon, U.~Von Luxburg, S. Bengio, H. Wallach, R. Fergus, S.
  Vishwanathan, and R. Garnett, editors, {\em NeurIPS}, 2017.

\bibitem{posec3d}
Haodong Duan, Yue Zhao, Kai Chen, Dahua Lin, and Bo Dai.
\newblock Revisiting skeleton-based action recognition.
\newblock In {\em CVPR}, 2022.

\bibitem{Finn16}
Chelsea Finn, Ian~J. Goodfellow, and Sergey Levine.
\newblock Unsupervised learning for physical interaction through video
  prediction.
\newblock In {\em NeurIPS}, 2016.

\bibitem{fox2021stylevideogan}
Gereon Fox, Ayush Tewari, Mohamed Elgharib, and Christian Theobalt.
\newblock Stylevideogan: A temporal generative model using a pretrained
  stylegan, 2021.

\bibitem{GAN}
Ian Goodfellow, Jean Pouget-Abadie, Mehdi Mirza, Bing Xu, David Warde-Farley,
  Sherjil Ozair, Aaron Courville, and Yoshua Bengio.
\newblock Generative adversarial nets.
\newblock In {\em NeurIPS}, 2014.

\bibitem{GrathwohlW16}
Will Grathwohl and Aaron Wilson.
\newblock Disentangling space and time in video with hierarchical variational
  auto-encoders.
\newblock {\em CoRR}, 2016.

\bibitem{Hess2}
Berk Hess.
\newblock Convergence of sampling in protein simulations.
\newblock 2002.

\bibitem{ho2022video}
Jonathan Ho, Tim Salimans, Alexey Gritsenko, William Chan, Mohammad Norouzi,
  and David~J Fleet.
\newblock Video diffusion models.
\newblock {\em arXiv:2204.03458}, 2022.

\bibitem{slomo18}
Huaizu Jiang, Deqing Sun, Varun Jampani, Ming{-}Hsuan Yang, Erik~G.
  Learned{-}Miller, and Jan Kautz.
\newblock Super slomo: High quality estimation of multiple intermediate frames
  for video interpolation.
\newblock In {\em CVPR}, 2018.

\bibitem{Karras2021}
Tero Karras, Miika Aittala, Samuli Laine, Erik H\"ark\"onen, Janne Hellsten,
  Jaakko Lehtinen, and Timo Aila.
\newblock Alias-free generative adversarial networks.
\newblock In {\em NeurIPS}, 2021.

\bibitem{Karras2019stylegan2}
Tero Karras, Samuli Laine, Miika Aittala, Janne Hellsten, Jaakko Lehtinen, and
  Timo Aila.
\newblock Analyzing and improving the image quality of {StyleGAN}.
\newblock In {\em CVPR}, 2020.

\bibitem{t2v}
Seyed~Mehran Kazemi, Rishab Goel, Sepehr Eghbali, Janahan Ramanan, Jaspreet
  Sahota, Sanjay Thakur, Stella Wu, Cathal Smyth, Pascal Poupart, and Marcus~A.
  Brubaker.
\newblock Time2vec: Learning a vector representation of time.
\newblock {\em CoRR}, 2019.

\bibitem{DlibmlAM}
Davis~E. King.
\newblock Dlib-ml: A machine learning toolkit.
\newblock {\em J. Mach. Learn. Res.}, 2009.

\bibitem{RAV}
Steven~R. Livingstone and Frank~A. Russo.
\newblock The ryerson audio-visual database of emotional speech and song
  (ravdess): A dynamic, multimodal set of facial and vocal expressions in north
  american english.
\newblock {\em PLOS ONE}, 13:1--35, 2018.

\bibitem{DMS16}
Michael Mathieu, Camille Couprie, and Yann Lecun.
\newblock Deep multi-scale video prediction beyond mean square error.
\newblock 2016.

\bibitem{miyato2018cgans}
Takeru Miyato and Masanori Koyama.
\newblock c{GAN}s with projection discriminator.
\newblock In {\em ICLR}, 2018.

\bibitem{TSGAN21_WACV}
Andres Munoz, Mohammadreza Zolfaghari, Max Argus, and Thomas Brox.
\newblock Temporal shift gan for large scale video generation.
\newblock In {\em WACV}, January 2021.

\bibitem{Niklaus_CVPR_2017}
Simon Niklaus, Long Mai, and Feng Liu.
\newblock Video frame interpolation via adaptive convolution.
\newblock In {\em CVPR}, 2017.

\bibitem{roich2021pivotal}
Daniel Roich, Ron Mokady, Amit~H Bermano, and Daniel Cohen-Or.
\newblock Pivotal tuning for latent-based editing of real images.
\newblock {\em ACM Trans. Graph.}, 2021.

\bibitem{TGAN2017}
Masaki Saito, Eiichi Matsumoto, and Shunta Saito.
\newblock Temporal generative adversarial nets with singular value clipping.
\newblock In {\em ICCV}, 2017.

\bibitem{TGAN2020}
Masaki Saito, Shunta Saito, Masanori Koyama, and Sosuke Kobayashi.
\newblock Train sparsely, generate densely: Memory-efficient unsupervised
  training of high-resolution temporal gan, 2020.

\bibitem{collpaseCon}
Mohamad Shahbazi, Martin Danelljan, Danda~Pani Paudel, and Luc~Van Gool.
\newblock Collapse by conditioning: Training class-conditional {GAN}s with
  limited data.
\newblock In {\em ICLR}, 2022.

\bibitem{singer2023makeavideo}
Uriel Singer, Adam Polyak, Thomas Hayes, Xi Yin, Jie An, Songyang Zhang, Qiyuan
  Hu, Harry Yang, Oron Ashual, Oran Gafni, Devi Parikh, Sonal Gupta, and Yaniv
  Taigman.
\newblock Make-a-video: Text-to-video generation without text-video data.
\newblock In {\em ICLR}, 2023.

\bibitem{Skorokhodov_2022_CVPR}
Ivan Skorokhodov, Sergey Tulyakov, and Mohamed Elhoseiny.
\newblock Stylegan-v: A continuous video generator with the price, image
  quality and perks of stylegan2.
\newblock In {\em CVPR}, 2022.

\bibitem{Songsriin2019FaceVG}
Kritaphat Songsri-in and Stefanos Zafeiriou.
\newblock Face video generation from a single image and landmarks.
\newblock {\em FG}, 2019.

\bibitem{tian2021a}
Yu Tian, Jian Ren, Menglei Chai, Kyle Olszewski, Xi Peng, Dimitris~N. Metaxas,
  and Sergey Tulyakov.
\newblock A good image generator is what you need for high-resolution video
  synthesis.
\newblock In {\em ICLR}, 2021.

\bibitem{tulyakov2018mocogan}
Sergey Tulyakov, Ming-Yu Liu, Xiaodong Yang, and Jan Kautz.
\newblock Mocogan: Decomposing motion and content for video generation.
\newblock In {\em CVPR}, 2018.

\bibitem{FVD18}
Thomas Unterthiner, Sjoerd van Steenkiste, Karol Kurach, Raphael Marinier,
  Marcin Michalski, and Sylvain Gelly.
\newblock Towards accurate generative models of video: A new metric \&
  challenges.
\newblock {\em arXiv preprint arXiv:1812.01717}, 2018.

\bibitem{VGAN16}
Carl Vondrick, Hamed Pirsiavash, and Antonio Torralba.
\newblock Generating videos with scene dynamics.
\newblock In {\em NeurIPS}, 2016.

\bibitem{WalkerDGH16}
Jacob Walker, Carl Doersch, Abhinav Gupta, and Martial Hebert.
\newblock An uncertain future: Forecasting from static images using variational
  autoencoders.
\newblock In {\em ECCV}, 2016.

\bibitem{MEAD20}
Kaisiyuan Wang, Qianyi Wu, Linsen Song, Zhuoqian Yang, Wayne Wu, Chen Qian, Ran
  He, Yu Qiao, and Chen~Change Loy.
\newblock Mead: A large-scale audio-visual dataset for emotional talking-face
  generation.
\newblock In {\em ECCV}, 2020.

\bibitem{wang2018vid2vid}
Ting-Chun Wang, Ming-Yu Liu, Jun-Yan Zhu, Guilin Liu, Andrew Tao, Jan Kautz,
  and Bryan Catanzaro.
\newblock Video-to-video synthesis.
\newblock In {\em NeurIPS}, 2018.

\bibitem{ImaGin20}
Yaohui WANG, Piotr Bilinski, Francois Bremond, and Antitza Dantcheva.
\newblock Imaginator: Conditional spatio-temporal gan for video generation.
\newblock In {\em WACV}, March 2020.

\bibitem{x2face}
Olivia Wiles, A.~Sophia Koepke, and Andrew Zisserman.
\newblock X2face: {A} network for controlling face generation using images,
  audio, and pose codes.
\newblock In {\em ECCV}, 2018.

\bibitem{xiang2020zooming}
Xiaoyu Xiang, Yapeng Tian, Yulun Zhang, Yun Fu, Jan~P. Allebach, and Chenliang
  Xu.
\newblock Zooming slow-mo: Fast and accurate one-stage space-time video
  super-resolution.
\newblock In {\em CVPR}, 2020.

\bibitem{coninvert21}
Yangyang Xu, Yong Du, Wenpeng Xiao, Xuemiao Xu, and Shengfeng He.
\newblock From continuity to editability: Inverting gans with consecutive
  images.
\newblock In {\em ICCV}, 2021.

\bibitem{PVDM}
Sihyun Yu, Kihyuk Sohn, Subin Kim, and Jinwoo Shin.
\newblock Video probabilistic diffusion models in projected latent space.
\newblock In {\em CVPR}, 2023.

\bibitem{Digan2022}
Sihyun Yu, Jihoon Tack, Sangwoo Mo, Hyunsu Kim, Junho Kim, Jung-Woo Ha, and
  Jinwoo Shin.
\newblock Generating videos with dynamics-aware implicit generative adversarial
  networks.
\newblock In {\em ICLR}, 2022.

\bibitem{DwNet}
Polina Zablotskaia, Aliaksandr Siarohin, Bo Zhao, and Leonid Sigal.
\newblock Dwnet: Dense warp-based network for pose-guided human video
  generation.
\newblock In {\em BMVC}, 2019.

\bibitem{zhang2018perceptual}
Richard Zhang, Phillip Isola, Alexei~A Efros, Eli Shechtman, and Oliver Wang.
\newblock The unreasonable effectiveness of deep features as a perceptual
  metric.
\newblock In {\em CVPR}, 2018.

\bibitem{thinplate22}
Jian Zhao and Hui Zhang.
\newblock Thin-plate spline motion model for image animation.
\newblock {\em CVPR}, 2022.

\end{thebibliography}
}

\end{document}